\documentclass{article}
\usepackage{ijcai18}
\usepackage{graphicx}
\usepackage{times}
\usepackage{xcolor}
\usepackage{soul}
\usepackage[utf8]{inputenc}
\usepackage[small]{caption}

\title{Multiple Target Tracking by Learning Feature Representation and Distance Metric Jointly}
\author{
Jun Xiang$^1$,
Guoshuai Zhang$^1$,
Jianhua Hou$^1$,
Sang Nong$^2$,
Rui Huang$^3$
\\
$^1$ South-Central University for Nationalities, Wuhan, China \\
$^2$ Huazhong University of Science and Technology,
Wuhan, China\\
$^3$ The Chinese University of Hong Kong, Shenzhen, China  \\
junxiang@hust.edu.cn,
zil@scuec.edu.cn,
 nsang@hust.edu.cn,
 ruihuang@cuhk.edu.cn,
}

\begin{document}
\maketitle
\begin{abstract}
  Designing a robust affinity model is the key issue in multiple target tracking (MTT). With the release of  more challenging benchmarks, the association performance based on traditional hand-crafted features is unsatisfactory. Although several current MTT methods adopt deep CNN features directly, they don’t resort to metric learning which can significantly improve the model’s discriminability. This paper proposes a novel affinity model by learning feature representation and distance metric jointly in a unified deep architecture. Specifically, we design a CNN network to obtain appearance cue tailored towards person Re-ID, and an LSTM network for motion cue to predict target position, respectively. Both cues are combined with a triplet loss function, which performs end-to-end learning of the fused features in a desired embedding space. Experiments in the challenging MOT benchmark demonstrate, that even by a simple Linear Assignment strategy fed with affinity scores of our method, very competitive results are achieved when compared with the most recent state-of-the-art approaches.
\end{abstract}

\section{Introduction}
Detection based multiple target tracking (MTT) has become increasingly popular thanks to the great improvements in object detection~\cite{viola2001rapid,dalal2005histograms,felzenszwalb2010object}. Given detection responses by a pre-trained detector, this paradigm consists of two key components: an affinity model for estimating the linking probability between detections or tracklets, and an optimization strategy for association. This paper focuses on the first issue.

Feature representation is critical to affinity model designing. In the last two decades, many efficient feature descriptors have been developed to construct a robust appearance model, including HOG features~\cite{dalal2005histograms}, local binary patterns~\cite{wang2009hog}, color histograms~\cite{nummiaro2003adaptive}, etc. Meanwhile, some works have combined other cues like motion and position with the target appearance to further encode the dynamics and interaction among targets~\cite{yang2014multi,xiang2016multitarget}. However, with the release of new standardized benchmarks like MOTChallenge~\cite{leal2015motchallenge,milan2016mot16}, more challenging scene is presented. For example, the variation between intra-target could be larger than that of extra-target due to great appearance deforming, very low illumination and severe occlusions. In this situation, tracking performance based on the above traditional hand-crafted features is unsatisfactory.

Recently, deep neural networks have been employed in multiple target tracking framework~\cite{kim2015multiple,leal2016learning,sadeghian2017tracking,tang2017multiple}. As a common practice, target representation is learned by convolutional neural networks (CNN), and is used directly to compute pairwise affinity measure (linking probability) which determines whether two detections belong to the same target. This paradigm seldom considers metric learning, which is widely used in the task of person re-identification (Re-ID) and could be exploited here to significantly improve the feature’s discriminability.

This paper aims to address two key issues in a unified deep architecture: integrating multiple cues into feature description, and jointly learning feature representation and distance metric. Specifically, we employ a CNN to extract appearance cue tailored towards person Re-ID, and a Long Short-Term Memory (LSTM) network to extract motion cue aiming at predicting target position. Finally, the CNN and LSTM are combined with  a triplet loss function, which performs end-to-end deep metric learning and generates the embeddings of the fused appearance and motion features. The proposed ``CNN+LSTM ''model together with a triplet loss function can be considered as learning a mapping function that maps each detection into an embedding space where the difference between detections of the same target is less than that of different targets. Therefore, we can simply compare a pair of detections by computing the Euclidean distance of their embeddings since we directly optimize the network for the task of feature and distance metric learning jointly.

Our main contributions are as follows:
\begin{itemize}
\item Within the context of multiple target tracking, we propose a unified deep architecture for associating detections, where feature representation and distance metric are learned jointly.
\item To introduce the mechanism for multiple cue based feature  representation, we design a CNN and an LSTM network to obtain appearance cue tailored towards person Re-ID and motion cue aiming at predicting target position, respectively.
\item A triplet loss function is introduced that not only integrates CNN and LSTM to perform end-to-end learning of multiple cues representations, but produces the desired embedding space, which renders an additional metric learning step obsolete.
\end{itemize}

\section{Related Work}
\subsubsection{Appearance model}
In tracking-by-detection paradigm, the critical component  is to design a robust affinity model to compute the pairwise affinity measure (i.e. linking probability) between two detections across time. Most existing methods adopt weak affinity measures based on appearance model such as spatial affinity, e.g. bounding box overlap or Euclidean distance~\cite{andriyenko2012discrete,pirsiavash2011globally}, or simple appearance similarity, e.g. intersection kernel with color histogram~\cite{zamir2012gmcp}.

A recent trend is to exploit the representation power of deep architecture and develop CNN-based appearance features to effectively model similarity between detections~\cite{leal2016learning,sadeghian2017tracking,tang2017multiple,wang2016joint}. A common manner in this trend is to train a Siamese CNN to discriminate whether two detections belong to the same target by using a binary classification loss. However, models trained with the verification loss, i.e. the softmax output, only answer the question ``How similar are these two detections?''~\cite{hermans2017defense}. Intuitively, the answer from a verification loss is ``arbitrary'' to some degree, since it considers neither ``where and when these detections originated'', nor other data points except the point pair being compared. To alleviate this problem, Leal-Taixe et al.~\cite{leal2016learning} train a CNN in a Siamese configuration, and the CNN output are then combined with contextual features by a gradient boosting algorithm.

Instead of binary classification, a more appropriate viewpoint is to treat MTT as a retrieval or Re-ID problem, and to build appearance model based on identity classification loss (identity preserving loss) . Along this line, Tang et al.~\cite{tang2017multiple} developed a Siamese ID-Net for person Re-ID, where the appearance learned by deep networks and body pose information are combined. While identity classification loss is used in training stage, a binary classification is adopted in the test time due to the unknown number of targets in MTT, thus falling into the verification  ``trap''.

This paper introduces a metric learning strategy, a widely used technique in person Re-ID and bipartite graph matching, to mitigate the above problem. We choose triplet loss function to map CNN features into an embedding space where distances for different appearances of the same target is smaller than that of different targets. The triplet-based embedding is more suitable for multiple target tracking where the data association can be viewed as a bipartite graph matching problem.
\subsubsection{Motion model}
 To improve the affinity model’s accuracy, motion models are presented and often combined with appearance to predict the target location~\cite{milan2017online,xiang2016hough,breitenstein2009robust,wang2016joint,sadeghian2017tracking}. Milan et al.~\cite{milan2017online} proposed a RNN-based network to learn complex motion models  under the framework of Bayesian fltering, and the temporal dynamics of targets learned by RNN is utilized to perform state prediction and updating as well as track management. Popular motion models used in MTT are often linear~\cite{xiang2016hough,breitenstein2009robust,wang2016joint} with a priori assumption that targets follow a linear movement with constant velocity across frames . Wang et al.~\cite{wang2016joint} constructs a Siamese CNN and a traditional linear model to extract appearance and motion features, respectively. The deep appearance features and hand-crafted motion features are integrated to estimate the linking probability for target association. However, the simple mechanism behind linear model makes it hard to produce a more accurate prediction in complex and crowded scene, leading to unrealistic or unreasonable trajectories due to the complexity of human motion patterns. To tackle this shortcoming, non-linear motion models are developed to capture more complex dynamics and motion dependency between targets. Sadeghian et al.~\cite{sadeghian2017tracking} presented an LSTM model to predict similar motion patterns by considering the past movements of an object and predicting its future trajectory. To encode long-term temporal dependencies, a hierarchical RNN is used to jointly reason on motion, appearance and interaction cues over a temporal window.

 Similar to~\cite{sadeghian2017tracking}, our approach integrates multiple cues into feature description and employs an LSTM network to model motion dependency between targets. However, our system differs in two aspects. First, instead of a regular Siamese CNN in~\cite{sadeghian2017tracking}, we employ a CNN tailored towards person Re-ID to extract appearance cue. Second, while multiple cues are merged by a RNN with binary classification loss in ~\cite{sadeghian2017tracking}, our approach adopts triplet loss to map the merged features into an embedding space, rendering a mechanism of distance metric learning.

\section{Multiple Target Tracking Framework}
\subsection{Architecture Overview}
We formulate multiple target tracking as a data association problem. Given a set of already tracked targets at current time, each newly detection in the next time should be associated uniquely to a corresponding target. To this end, an affinity model outputs linking probability (or linking cost) to construct a bipartite graph between already tracked targets and newly detections, and Hungarian algorithm is utilized to perform associations across time.

\begin{figure}
	\centering
		\includegraphics[width=80mm]{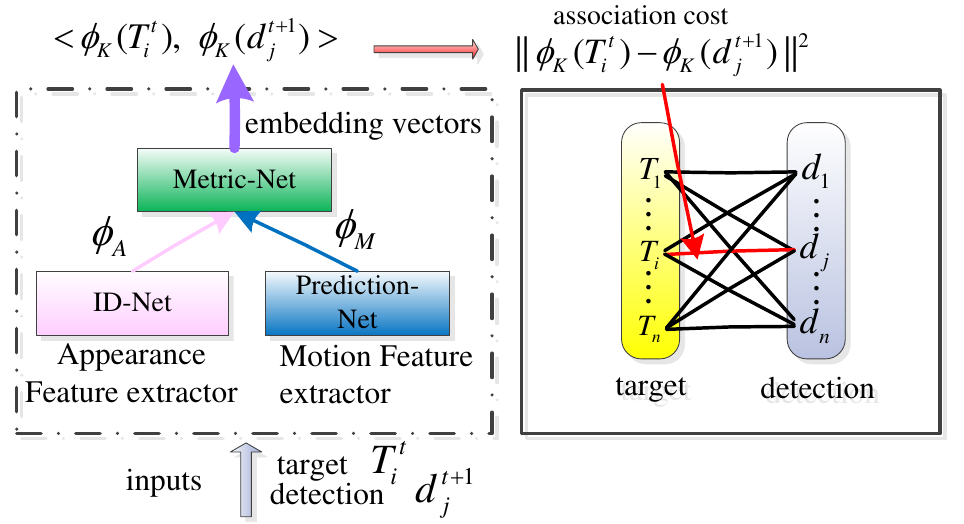}
	\caption{\label{fig:Fig1} Architecture overview. In affinity model (the dashed rectangle), the ID-Net and Prediction-Net extract appearance cue and motion cue, respectively. Both cues are combined in Metric-Net with a triplet loss function. The Metric-Net performs end-to-end learning of multiple cue representations and produces the desired embedding features. In matching part (the soiled rectangle), a bipartite graph is constructed by association cost between already tracked targets and newly detections, and matching is achieved by Hungarian algorithm.
	}
\end{figure}

The core task of this paper is to present a novel detection affinity model by learning feature representation and distance metric jointly in a unified deep architecture. The overview of our framework is shown in Fig.~\ref{fig:Fig1}.  The proposed affinity model consists of three components. The first is a CNN network tailored towards person Re-ID for appearance, denoted as ID-Net. The second is an LSTM network used to predict target position, denoted as Prediction-Net. The output of the both networks, denoted by ${\phi _A}$ and ${\phi _M}$,  are combined in the third part equipped with a triplet loss function, namely Metric-Net. The Metric-Net performs end-to-end learning of multiple cue representations, and outputs  embedding feature ${\phi _K}(T_i^t)$ and ${\phi _K}(d_j^{t + 1})$, which are used to output the association cost between $T_i^t$ and $d_j^{t+1}$. In the rest of this section, we describe each component of our method.
\subsection{Appearance model}
The motivation of our appearance model is to develop more discriminative and robust  representations for visual feature property. To this end, we employ a CNN with an identity classification loss, that is usually used in person Re-ID.
\subsubsection{Data Collection}
It is well known that deep architectures require vast amounts of training data in order to avoid overfitting of model. Training images are collected from the MOT15 benchmark training set~\cite{leal2015motchallenge} and 5 sequences in the MOT16 benchmark training set~\cite{milan2016mot16}. We also collect person identity examples from the CUHK03~\cite{li2014deepreid}  and Market-1501~\cite{li2014deepreid} datasets. For validation set, we use the MOT16-02 and MOT16-11 sequences from the MOT16 training set . Overall a total of 2551 identities are used for training and 123 identities for validating.
\subsubsection{Architectures}
Similar to~\cite{tang2017multiple}, we use VGG-16 Net~\cite{simonyan2014very} as the base CNN architecture. Specifically, by training VGG-16 to recognize $Y = 2551$ unique identities, the learning can be viewed as a $Y$-way classification problem. Training images are re-sized to $112 \times 224 \times 3$. Each image $I_i$  associates to a ground truth identity label ${l_i} \in \{ 1,2 \cdots Y\}$. The network is trained by the softmax loss to estimate the probability of each image being each label by a forward pass. Note that in test time, we remove the final softmax layer and use the fully connected layer ${\Phi _{f7}}$ (4096-dimition) as the appearance feature ${\phi _A}({I_i})$.
\subsection{Motion model}
\begin{figure}
	\centering
		\includegraphics[width=80mm]{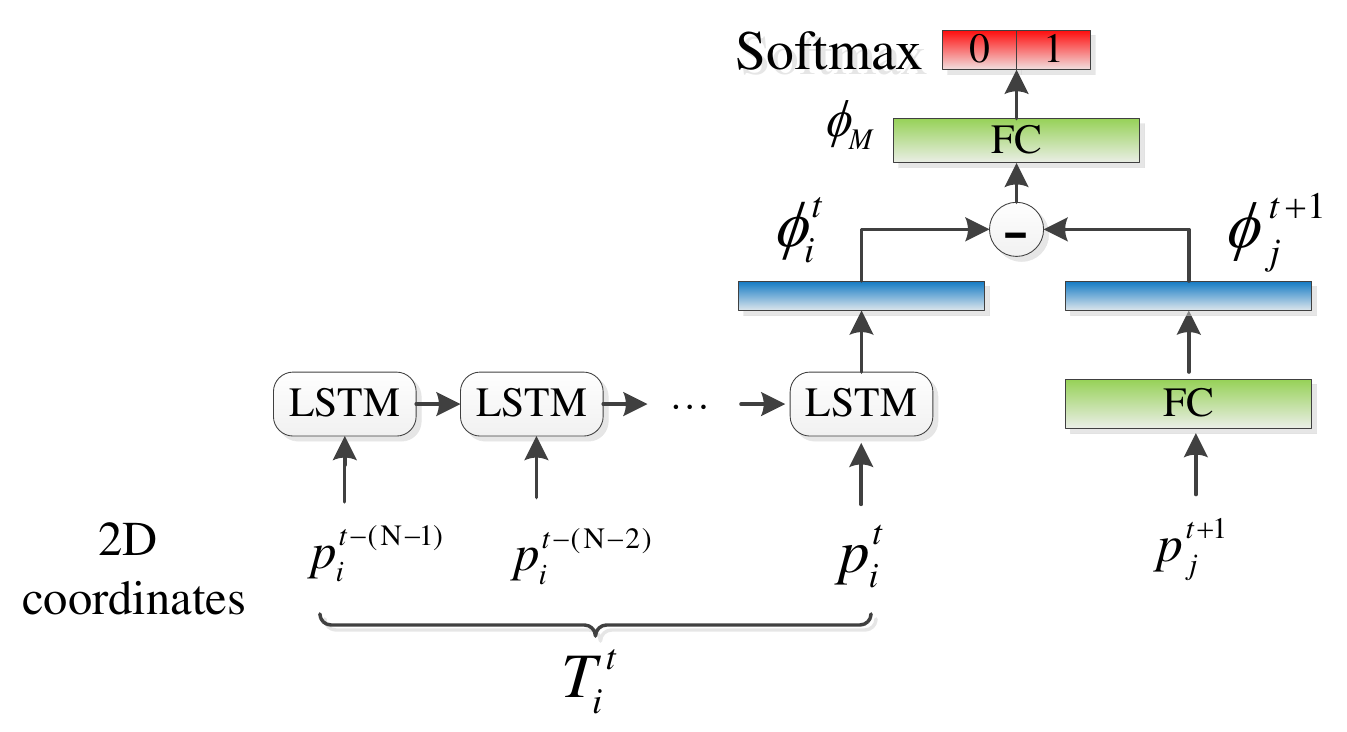}
	\caption{\label{fig:Fig2} Prediction-Net model.  The inputs are 2D coordinates.
	}
\end{figure}

Intuitively, the dynamics over trajectories of a specific target encodes its motion pattern that can be exploited to predict future position for a target. In this paper, we explore a Long Short-Term Memory (LSTM) to model target motions over positions(inspired by~\cite{sadeghian2017tracking}). The proposed Prediction-Net can help re-identifying targets that are occluded or lost, since it provides a prediction the target might be located. Our Prediction-Net model is illustrated in Fig.~\ref{fig:Fig2}.
\subsubsection{Architecture}
The task of  motion model is to determine whether a trajectory should be located at a particular position or not. Our LSTM accepts as inputs the positions of trajectory $i$ ( $N$ frames in length),  denoted as $T_i^t = [p_i^{t - (N - 1)},p_i^{t - (N - 2)}, \cdots ,p_i^t]$, and produces a $H$-dimensional output $\phi _i^t$ . We also pass a position $p_j^{t + 1}$ from trajectory $j$ for time $t+1$   (which we wish to determine whether it corresponds to the true trajectory $T_i^t$ or not) through a fully-connected layer that maps it to a  $H$-dimensional vector, denoted as ${\phi _j}^{t + 1}$ . The difference between $\phi _i^t$ and ${\phi _j}^{t + 1}$ is passed to another fully connected layer, followed by a softmax layer to produce a probability estimation over binary classification. In test stage, we remove the final softmax layer and the output of the top FC layer is used as the final motion vector, denoted as ${\phi _M}(T_i^t,p_j^{t + 1})$ .
\subsubsection{Data Collection}
Due to the very tedious and time-consuming task of video annotation, only very limited amount of real data for pedestrian tracking is publicly available today. We therefore resort to synthetic data augmentation as in~\cite{milan2017online}, by sampling from a simple generative trajectory model learned from MOT15 and MOT16. We refer to~\cite{milan2017online} for more details.
There are about 100K trajectories in the collected training set, each of 20 frames in length. The data is divided into mini-batches of 10 samples per batch and normalized to the range [-0.5, 0.5], w.r.t. the image dimensions. As mentioned previously, each sample is a data pair consisting of a trajectories $T_i^t$  ($N$ frames in length) and a position $p_j^{t + 1}$ . While positive samples are generated by randomly sampling $T_i^t$  and its true position $p_i^{t{\rm{ + }}1}$ , negative samples consists of $T_i^t$  and a position $p_j^{t{\rm{ + }}1}$ from a different trajectory $j$.
\subsection{Metric Learning}
\begin{figure}
	\centering
		\includegraphics[width=85mm]{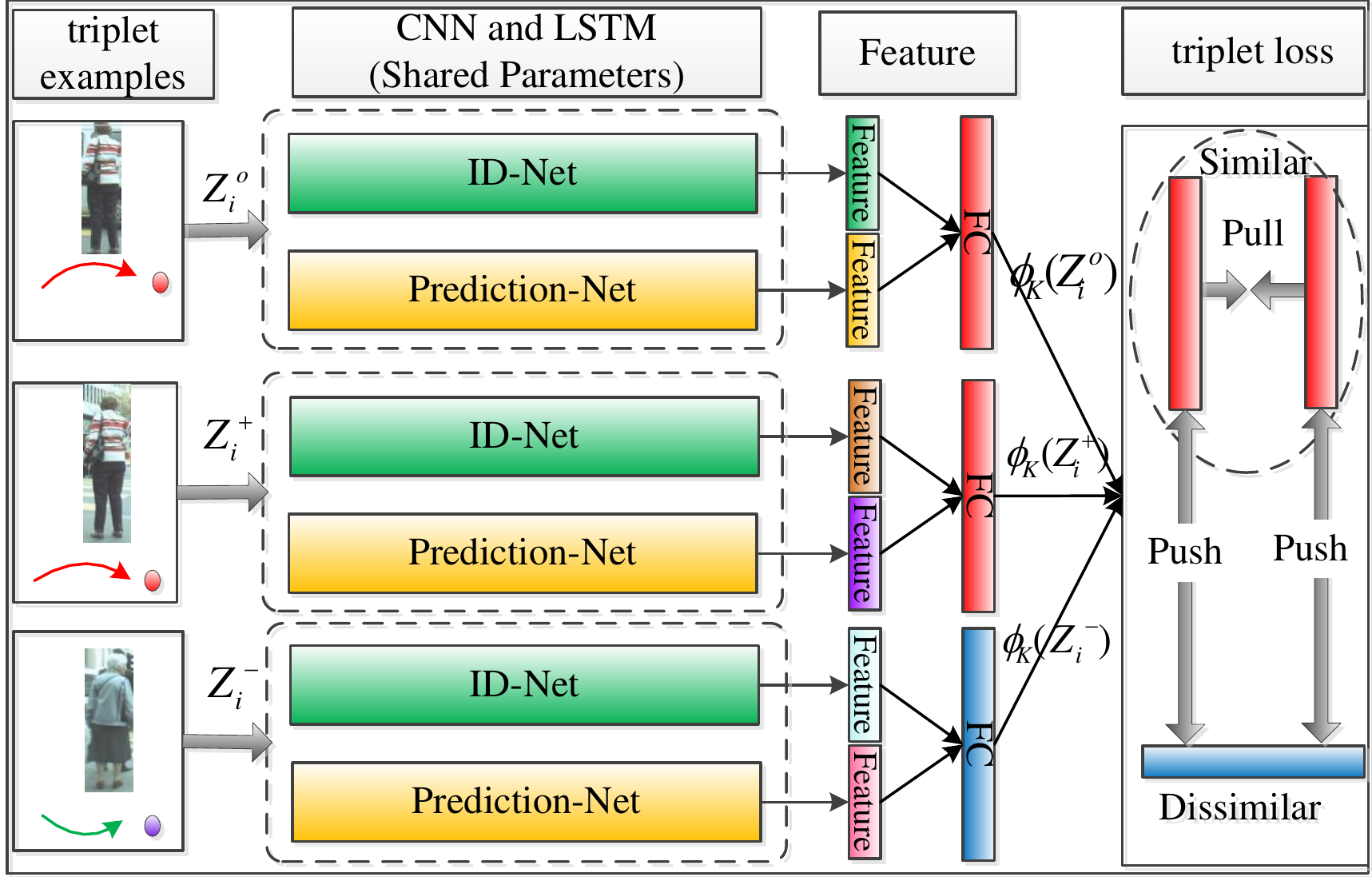}
	\caption{\label{fig:Fig3} Metric-Net training framework. Triplet training images are fed into three-channel CNN-LSTM networks with the shared parameter set. The triplet loss function is used to train the Metric-Net, making the distance between inputs of the same targets is less than that of different targets.
	}
\end{figure}

 In this paper, our metric learning encodes dependencies across appearance and motion automatically by using a triplet loss function as illustrated in Fig.~\ref{fig:Fig3},  denoted as Metric-Net. Firstly, we pre-trained appearance and motion model separately, i.e, ID-Net and Prediction-Net. Then we train our Metric-Net with fine-tuning the weights of  individual components in an end-to-end fashion by fitting appearance and motion features into the triplet loss function. In the remaining of this section, we describe the detail for metric learning.
\subsubsection{Data Collection}
The components of triplet training examples come from image $I_i$  in ID-Net, as well as  trajectories $T_i^{t}$ and position $p_j^{t + 1}$ in Metric-Net. Inspired by the works in~\cite{schroff2015facenet,Cheng2016Person}, we construct Triplet example(also called instance)  ${Z_i} =  < Z_i^o,\;Z_i^ + ,\;Z_i^ -  > $  with three items. In anchor $Z_i^o = \{ {I_i},T_i^{ {t_1}},p_i^{ {t_1} + 1}\}$ , both $T_i^{ {t_1}}$ and $p_i^{ {t_1} + 1}$  are from a specific target $i$ , and  ${I_i}$ is a randomly sampled detection image from $T_i^{t_1}$ . Anchor-positive $Z_i^ +  = \{ {I'_i},T_i^{ {t_2}},p_i^{ {t_2} + 1}\} $ is similar to anchor but with different time stamp $t_2$ . The underlying principle behind our metric learning is to pull together samples from the same class in terms of appearance and motion, while push apart those with either different class in the terms of appearance or unreasonable motion state. Consequently, the trajectory and position in Anchor-negative $Z_i^ -  = \{ {I_j},T_j^{  {t_3}},p_k^{  {t_3} + 1}\}$ come from a different target  $j$ (i.e. $i \ne j$ ) . Note that in this case we don’t care about whether  $p_k^{t_3 + 1}$ is the real position of  $T_j^{t_3}$ or not.

For experiments, we collect triplet examples from MOT15 benchmark training set and 6 sequences of the MOT16 benchmark training set . We use the MOT16-02 sequences from the MOT16 training set as test sets. Overall a total of 851 identities are used for training and 54 identities for testing. We generate the triplets as follows: For each batch of 100 instances, we select 5 persons and generate 20 instances for each person. In each triplet instance, the anchor and anchor-positive are randomly selected from the same identity, and the negative one is also randomly selected, but from the remaining identities.
\subsubsection{Architecture}
To train Metric-Net, we design a three-channel CNN-LSTM Model with the shared parameter set. See Fig.~\ref{fig:Fig3}.  In each channel, one item in triplet training example ${Z_i}$  is mapped into a learned feature space to form a $4096 + H$ dimensional vector by concatenating CNN and LSTM features. A subsequent FC layer is employed for each channel which brings this concatenated feature to a $K=256$ dimensional embedding space by a triplet loss function, where the embedding feature of ${Z_i}$  is represented by ${\phi _K}({Z_i}) =  < {\phi _K}(Z_i^o), {\phi _K}(Z_i^ + ), {\phi _K}(Z_i^ - ) > $ . The learned embedding space has the desirable property that the distance between  ${\phi _K}(Z_i^o)$ and ${\phi _K}(Z_i^ + )$  is less than the distance between ${\phi _K}(Z_i^o)$ and ${\phi _K}(\;Z_i^ - )$ by a predefined margin $\tau $, as described by the following equation
\begin{equation}
 d({\phi _K}(Z_i^o),\;{\phi _K}(Z_i^ + )) - d({\phi _K}(Z_i^ + ),{\phi _K}(\;Z_i^ - )) \le \tau
\end{equation}
where  $\tau $ is negative.

\section{Experiments}
In this section, we first describe implementation details and evaluation metrics. The performance of each component in our framework as well as the result of ablation study are then analyzed. Finally, we demonstrate validity of the proposed method by comparing with several recent state-of-the-art approaches on the benchmark of  MOTChallenge.

\subsection{Implementation Details}
To learn the ID-Net, our VGG model is pre-trained on the ImageNet Classification task and fine-turned with the MOT and person identity dataset. The learning rate is set initially to 0.0001 and  decreased by 10\% every 10 000 iterations. We set the maximum number of iterations to 600 000, which is enough to reach convergence.

Similar to~\cite{milan2017online}, we train Predition-Net from scratch. The weights are initialized from zero-mean Gaussian distributions with the standard deviations 0.01. The bias terms are set to 0. We use RMSprop ~\cite{tieleman2012rmsprop} to minimize the loss. The LSTM is trained with one layer and 300 hidden units. And the iteration step $N$ is experimentally set to 6 for all data sets. The learning rate is set initially to 0.0003 and decreased by 5\% every 20 000 iterations. We set the maximum number of iterations to 200 000, which is enough to reach convergence.

We use our ID-Net and Prediction-Net as initialization for learning Metric-Net, which makes the training faster and produces better results. In our experiments, the parameter margin of triplet loss function $\tau $ is set to $- 2$.

Now we describe data inputting strategy. For target $T_i^{t} = [p_i^{t - N},p_i^{t - (N - 1)}, \cdots ,p_i^t]$ , we pass its final detection image (at time $t$) to ID-Net, and pass the entire $T_j^t$  to Prediction-Net, where the first $N$ entries are inputted into the LSTM and the last entry into the FC layer . The output of our affinity model is the embedding feature ${\phi _K}(T_i^t)$. To produce  ${\phi _K}(d_j^{t+1})$, the detection image (at time ${t+1}$) of $d_j^{t + 1} $  is fed into ID-Net, and we pass $T_j^t$  and $d_j^{t + 1}$ to LSTM as described in sub-section 3.3.
\subsection{Evaluation metrics}
We follow the standard MOT2D Benchmark challenge~\cite{milan2016mot16} for evaluating multiple targets tracking performance. The metrics includes: Multiple Object Tracking Accuracy (MOTA$ \uparrow$), Multiple Object Tracking Precision (MOTP$ \uparrow $), Mostly Track targets (MT$ \uparrow $), Mostly Lost targets (ML$\downarrow$), False Positives (FP$ \downarrow $), False Negatives (FN$ \downarrow $), Fragmentation (FM$\downarrow $) and finally ID Switches (IDS$ \downarrow $). For items with ($\uparrow$), higher scores indicate better results; for those with ($ \downarrow $), lower scores indicate better results.
\subsection{ Experimental Analysis}
In this sub-section, we analyze the performance of each component of our model. We conduct experiments on MOT16 to investigate the validity of multiple cues and metric embeddings. 123 person identities collected from MOT16-02 and MOT16-11 are used as test samples. Detections that are considered as true positives for a certain identity are those whose intersection-over-union with the ground truth of the identity are larger than 0.5 .
\subsubsection{Validity of ID-Net}
We evaluate our appearance model for identity verification task. Given the true positive detections for all the test identities, we randomly select 2000 positive pairs assigned to the same identity, and 4000 negative pairs assigned to different identities as our test set. We use the verification accuracy metric, the ratio of correctly classified pairs. The verification result is obtained by comparing the  ${L_2}$  distance between the extracted features and a threshold. The threshold is obtained on a separate validation dataset by maximizing the verification accuracy, which is set to 0.5 in experiment. We also report the verification result of our ID-Net in a Siamese architecture manner (denoted as Siamese ID-Net), i.e. an additional FC layer on the top of the twin ID-Net is employed to model a 2-way classification.

It can be seen from Table.~\ref{tab:addlabel} that our ID-Net already produces reasonable verification accuracy. The performance is further improved by Siamese ID-Net, from 78.4\% to 84.2\%. Moreover, we also report Tracking Accuracy (MOTA) of the both networks on MOT-02. While the MOTA result is unsatisfactory due to considering no motion cue, it demonstrates that the ID-Net alone can extract meaningful appearance representation for association task. In addition, the result that ID-Net achieves a good verification accuracy but a poor racking accuracy has verifies our previous viewpoint, namely that models trained with the verification loss is ``arbitrary'' to some degree when applied in assignment task.
\begin{table}[t]
\caption{Validity of ID-Net}
\label{tab:addlabel}
\centering
\begin{tabular}{l|c|c }
\hline
Model & Verification Accuracy$\uparrow $  &MOTA$\uparrow $    \\
\hline
  ID-Net & 78.40\% & 10.4\% \\
  SID-Net & 84.20\% & 16.2\% \\
\hline
\end{tabular}
\end{table}
\subsubsection{Validity of Prediction-Net}
One of the hyper parameters of  Perdition-Net is the sequence length $N$ , which is the number of unrolled time steps used while training the LSTM model and enables Perdition-Net being capable to memorize long term dependencies of position cues across time. In this section, we investigate the impact of this parameter.
\begin{figure}
	\centering
		\includegraphics[width=80mm]{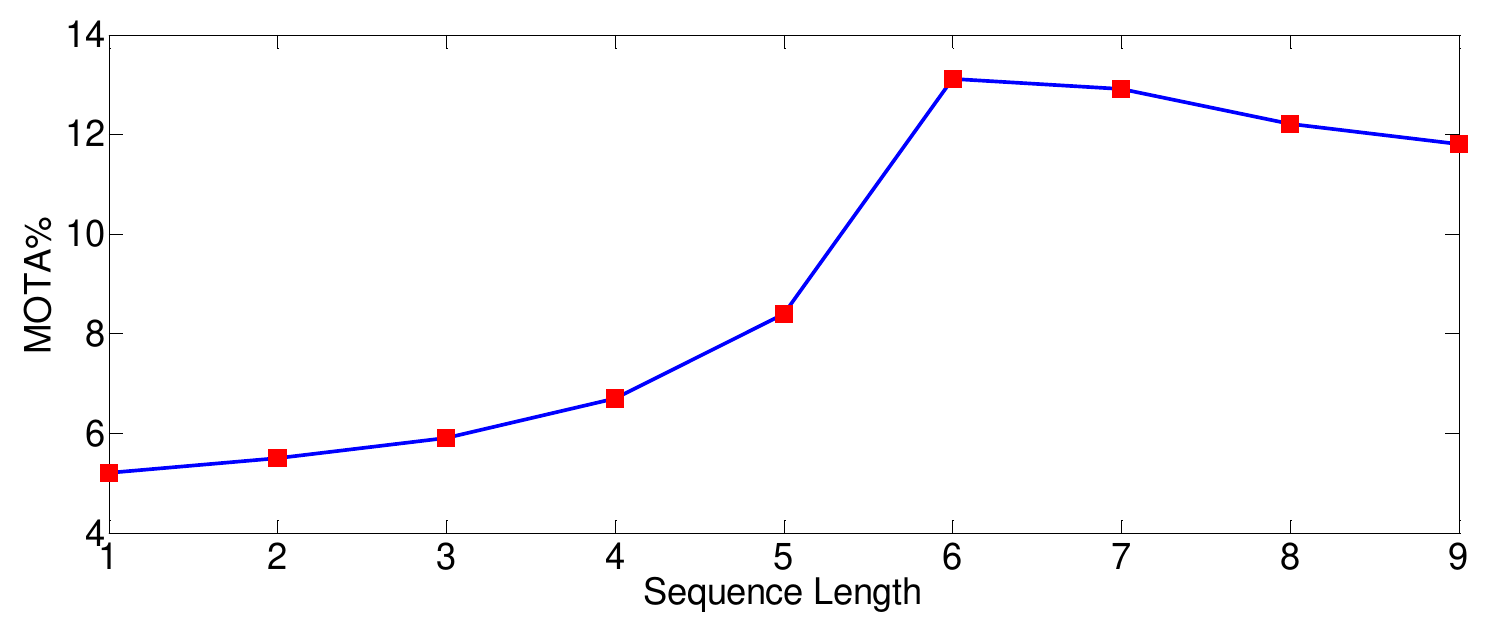}
	\caption{\label{fig:Fig4} Analysis of the used sequence length for our model on the MOT-02 set
	}
\end{figure}
 Fig.~\ref{fig:Fig4} shows the MOTA score on MOT-02 data set under different sequence length for our LSTM model. We can see that increasing the sequence length positively impacts the MOTA and the performance saturates after 6 frames, thus showing validity of our motion model. The result also confirms that an LSTM networks modeling dynamics over trajectories can be employed for tracking.
 \subsubsection{Ablation study}
We investigate the contribution of different components of our framework with tracking metrics on MOT-02 datase. To reveal the contribution of triplet loss, we conduct experiments for the same deep architectures with the verification loss. The evaluation results are presented in  Table.~\ref{tab:addlabe2} and summarized as follow:
\begin{itemize}
\item The appearance cue is the most important one. It can be explained by the fact that representations of people appearance can be learned for varying viewpoint and motion, while less easy to achieve by motion models, especially for monocular video sequences due to the complexity of motion.(Note that similar conclusions also are reported in~\cite{tang2017multiple,sadeghian2017tracking})
\item The motion cue helps to increase the performance. In highly crowded scenes with clutter and occlusions, our LSTM based motion model can facilitate localization of the targets, while appearance is usually sensitive since the observation likelihood of occluded targets may decrease drastically. In this case, both cues are complementary to make a better performance.
\item The triplet loss outperforms the verification loss by a large margin on the available datasets(from 17\% to 23\%). These results echo our claim that using triplet loss to optimize the embedding space is more suitable for retrieval or assignment task.
\end{itemize}
\begin{table}[htbp]
  \centering
  \caption{Analysis of our framework using different set of components. (A) Appearance, (M) Motion, (T) Triplet loss, and (V) Verification loss.}
    \resizebox{85mm}{11mm}{
    \begin{tabular}{cccccccc}
  \hline
    Tracker & MOTA$\uparrow$  & MOTP$\uparrow$& ML$\downarrow$    & MT$\uparrow$   & FP$\downarrow$    & FN$\downarrow$    & IDS $\downarrow$\\
  \hline
    A+T   & 21.90\% & 74.10\% & \textbf{63.00\%} & \textbf{11.10\%} & 246   & 13642 & 62 \\
    V+T   & 18.5 \% & 74.2  \% & 65.00\% & 7.40\% & 513  & 13944  & 73 \\
    A+M+T & \textbf{23.00\%} & 74.00\% & \textbf{63.00\%} & \textbf{11.10\%} & 188   & \textbf{13542} & \textbf{53} \\
  \hline
    A+V   & 16.2 \%      & 74.7\%      & 67.00\%     & 5.6\%     & 247     & 14555     & 148 \\
    M+V   & 13.10\% & \textbf{75.20\%} & 67.00\% & 7.40\% & 402   & 14679 & 410 \\
    A+M+V & 17.0\%      & 74.9\%     & 67.00\%     & 5.6\%     & \textbf{94}     & 14579     & 123 \\
  \hline
    \end{tabular}}%
  \label{tab:addlabe2}%
\end{table}%
\subsubsection{Comparison with the state of the art}
\begin{table*}[htbp]
  \centering
  \caption{Results on the MOT16 test dataset.Best in bold, our method is denoted by TripT}
\resizebox{135mm}{18mm}{
    \begin{tabular}{c|cccccccc}
 \hline
    Tracker & MOTA$\uparrow $  & MOTP$\uparrow $  & MT$\uparrow $   & ML$\downarrow$    & FP$\downarrow$    & FN$\downarrow$    & IDS$\downarrow$   & HZ$\uparrow $ \\
 \hline
    LMP~\cite{tang2017multiple}   &\textbf{48.8}  & \textbf{79}    & \textbf{18.20\%} & \textbf{40.10\%} & 6654  & \textbf{86245} & 481   & 0.5 \\
    AMIR~\cite{sadeghian2017tracking}  & 47.2  & 75.8  & 14.00\% & 41.60\% & \textbf{2681}  & 92856 & 774   & 1 \\
 \hline
    \textbf{TripT(Ours)} & 44.3     & 74.3     & 12.5\%     & 46.5\%     & 2797     & 98332     & 469     & 0.6 \\
  \hline
    Quad-CNN~\cite{son2017multi} & 44.1  & 76.4  & 14.60\% & 44.90\% & 6,388 & 94,775 & 745   & 1.8 \\
    oICF~\cite{kieritz2016online}  & 43.2  & 74.3  & 11.30\% & 48.50\% & 6,651 & 96,515 & 381   & 0.4 \\
    MHT-DAM~\cite{kim2015multiple} & 42.9  & 76.6  & 13.60\% & 46.90\% & 5,668 & 97,919 & 499   & 0.8 \\
    LINF1~\cite{fagot2016improving} & 41    & 74.8  & 11.60\% & 51.30\% & 7,896 & 99,224 & 430   & 1.1 \\
    EAMTT-pub~\cite{sanchez2016online} & 38.8  & 75.1  & 7.90\% & 49.10\% & 8,114 & 102,452 & 965   & 11.8 \\
    OVBT~\cite{ban2016tracking}  & 38.4  & 75.4  & 7.50\% & 47.30\% & 11,517 & 99,463 & 1,321 & 0.3 \\
    JPDA-m~\cite{rezatofighi2015joint} & 26.2  & 76.3  & 4.10\% & 67.50\% & 3,689 & 130,549 & \textbf{365}   & \textbf{22.2} \\
 \hline
    \end{tabular}}%
  \label{tab:addlabe3}%
\end{table*}%
We compare our method denoted by TripT with the best published results on the MOT16 test set. The quantitative results are presented in Table 3, and our method is among the top performing trackers. AMIR~\cite{sadeghian2017tracking} is the most relevant approach to ours, where appearance, motion and interaction cues are merged for affinity model and Hungarian algorithm is used for association. We believe our results can be further improved by utilizing other cues such as interaction in AMIR or robust optical flow for motion model. Quad-CNN~\cite{son2017multi} also adopted the metric learning strategy with the Quadruplet loss, and our performance is very close to Quad-CNN with 0.2 improvement in MOTA. In LMP~\cite{tang2017multiple}, more complex graph-cut association strategy is employed  to model temporal and structural correlation among targets, which could be utilized to improve tracking accuracy significantly.

\section{Conclusion}
In this work, we have presented a novel affinity model for data association under a unified deep architecture, where multiple cue feature representation and distance metric are jointly learned in an end-to-end fashion. Experiments in the challenging MOT benchmark show, that even employing a simple linear program algorithm for association, the proposed affinity model yields very competitive results compared with the most recent state-of-the-art approaches. In future, we believe by merging other cues for feature representation and more effective but often sophisticated optimization strategy for association,  better tracking results should be achieved.

\appendix
\bibliographystyle{named}
\bibliography{main}
\end{document}